\documentclass[pdflatex,sn-basic]{sn-jnl}

\usepackage{graphicx}%
\usepackage{multirow}%
\usepackage{amsmath,amssymb,amsfonts}%
\usepackage{amsthm}%
\usepackage{mathrsfs}%
\usepackage[title]{appendix}%
\usepackage{xcolor}%
\usepackage{textcomp}%
\usepackage{manyfoot}%
\usepackage{booktabs}%
\usepackage{algorithm}%
\usepackage{algorithmicx}%
\usepackage{algpseudocode}%
\usepackage{listings}%

\usepackage{setspace}
\usepackage{hyperref}
\usepackage{natbib}
\usepackage{multirow}

\usepackage{array}
\usepackage{tabularx}
\usepackage{longtable}
\usepackage{lineno}

\theoremstyle{thmstyleone}%
\theoremstyle{thmstyletwo}%

\theoremstyle{thmstylethree}%

\begin{document}

\footnotetext{Recently Accepted for publication in Wiley Interdisciplinary Reviews: Data Mining and Knowledge Discovery © 2026 Wiley Periodicals, Inc. All rights reserved. This version of the article has been accepted, after peer review but is not the version of record. The final version will be available at: https://doi.org/10.1002/widm.70093. Paper list at Github:https://github.com/sunxiaobei/awesome-gnn-based-link-prediction. This work is subject to a 12-month post-publication embargo.}

\title[A Survey on GNN-based Link Prediction: Techniques, Applications, and Challenges]{A Survey on GNN-based Link Prediction: Techniques, Applications, and Challenges}


\author[1,2]{Chengcheng Sun}
\author*[3]{Yajie Song}
\email{songyajie@cumt.edu.cn}
\author*[1,2]{Cheng Zhai}
\email{greatzc@cumt.edu.cn}
\author[3]{Jiayun Tian}
\author[3]{Jia Yang}
\author[3]{Xiaobin Rui}
\author[3]{Jian Zhang}
\author*[3]{Zhixiao Wang}
\email{zhxwang@cumt.edu.cn}
\author[4]{Philip S. Yu}

\affil[1]{\orgdiv{School of Safety Engineering}, \orgname{China University of Mining and Technology}, \orgaddress{Xuzhou, Jiangsu, \postcode{221116}, \country{China}}}

\affil[2]{\orgdiv{State Key Laboratory of Coal Mine Disaster Prevention and Control}, \orgname{China University of Mining and Technology}, \orgaddress{Xuzhou, Jiangsu, \postcode{221116}, \country{China}}}

\affil[3]{\orgdiv{School of Computer Science and Technology}, \orgname{China University of Mining and Technology}, \orgaddress{\street{Xuzhou}, \city{Jiangsu}, \postcode{221116}, \country{China}}}

\affil[4]{\orgdiv{Department of Computer Science}, \orgname{University of Illinois at Chicago}, \orgaddress{Chicago, IL, \country{USA}}}


\abstract{Graph Neural Networks (GNNs) have emerged as the leading paradigm for link prediction, enabling the inference of missing connections and the anticipation of potential future links. However, existing reviews lack systematic exploration specifically targeting underlying GNN architectures and diverse graph structures. To address this critical gap, this paper provides a comprehensive review of GNN-based link prediction from a novel and dedicated GNN perspective. We propose an innovative taxonomy that categorizes recent advancements based on techniques and applications. From a technique perspective, we focus on key GNN encoder architectures, including GCN-based, GAE-based, GAT-based, and GFormer-based methods, discussing their strengths and limitations. From an application perspective, we highlight prominent use cases of link prediction in knowledge graphs and recommendation systems, demonstrating their real-world impact. In addition, we examine the current challenges and discuss promising future directions.}




\maketitle

\section{Introduction}\label{sec1}

Graphs are essential for modeling complex systems due to their ability to represent intricate relationships. Link prediction is a fundamental task in such graphs, aiming to infer missing or future potential links between nodes based on the topology of the existing network. Typical applications include friend recommendation in social networks\cite{1apply_friendRec_2022}, product recommendation in e-commerce networks \cite{2apply_e-commerce_2013}, citation relationship prediction in citation networks \cite{3apply_citation_2009}, and movie recommendation in recommendation systems \cite{4apply_movieRec_2024}.

Traditionally, heuristic-based methods and shallow embedding-based methods are two classic approaches to link prediction. Heuristic-based methods primarily focus on exploring the connection patterns in a large amount of graph-structured data and designing specific heuristic formulas to predict new connections. For example, common neighbors (CNs) \cite{1tradition_CN_2001} is a popular heuristic that assumes node pairs sharing more neighbors are more likely to form connections. Similarly, the Jaccard index \cite{2tradition_Jaccard_1901}, the Adamic–Adar index \cite{6tradition_AA_2003}, and the Resource Allocation index \cite{5tradition_RA_2010} are all based on the assumption that node similarity can be inferred from their neighborhood information, and that such similarity indicates the likelihood of a connection between nodes. In contrast, shallow embedding-based methods aim to map nodes into lower-dimensional vectors while preserving their neighborhood structures and similar feature relationships within the graph. DeepWalk \cite{3tradition_DeepWalk_2014} and Node2vec \cite{4tradition_Node2vec_2016} treat nodes as words and sequences of connected nodes as sentences for node embedding. Other embedding methods, such as matrix factorization \cite{7tradition_MF_2011}, FSSDNMF \cite{8tradition_FSSDNMF_2022}, predict potential connections between nodes in a network by factorizing the adjacency matrix into two low-rank factor matrices. However, extending these methods to large-scale graphs may encounter limitations.

\begin{figure*}[!htbp]
\centerline{\includegraphics[width=\textwidth]{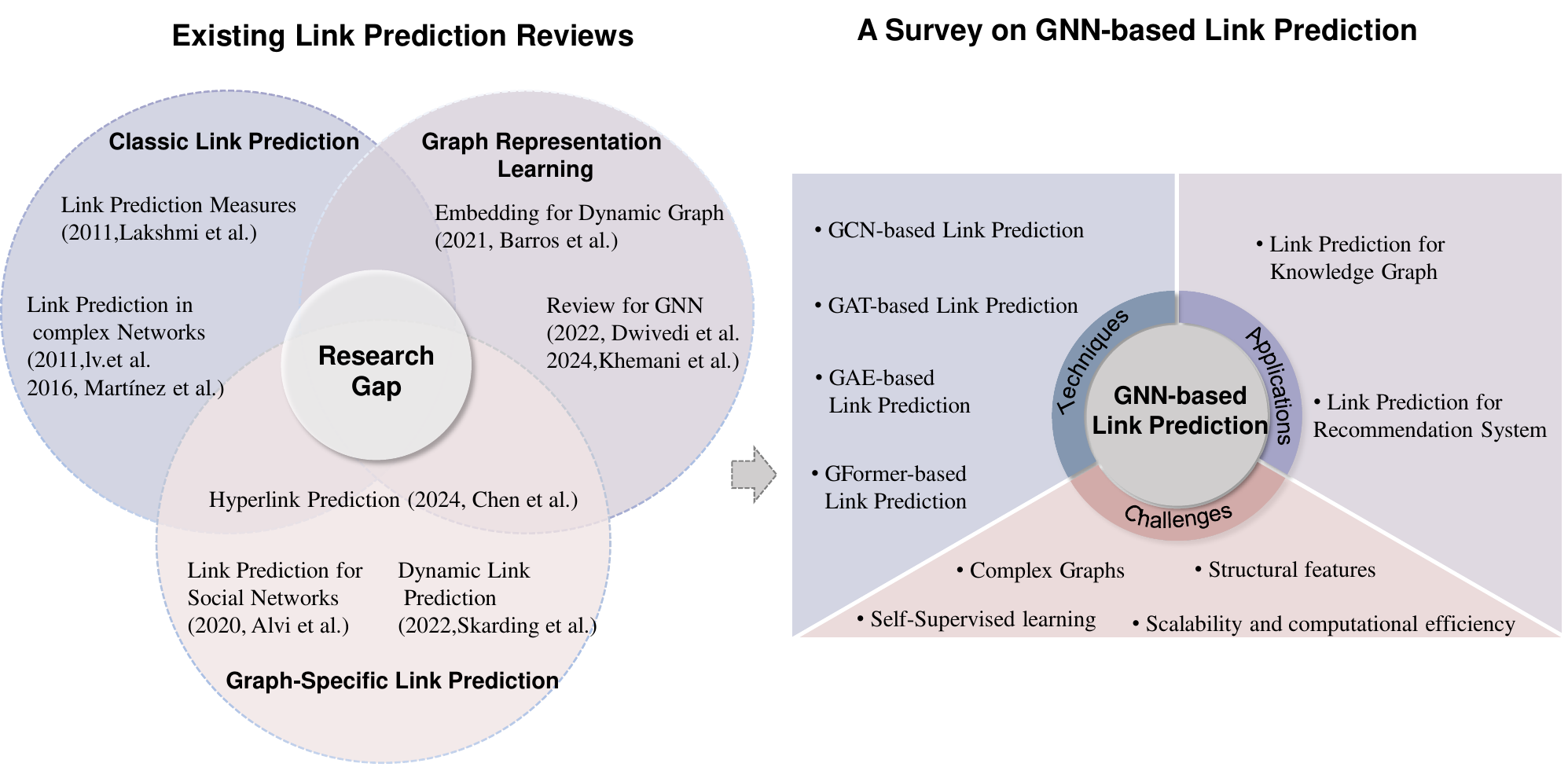}}
\caption{Motivation and Contributions: From Fragmented Perspectives to a Unified GNN-based Link Prediction Framework. The left panel identifies a critical research gap at the intersection of classic link prediction, graph representation learning, and graph-specific link prediction, where existing reviews lack a systematic exploration specifically targeting underlying GNN architectures and diverse graph structures. To address this critical gap, this survey provides a comprehensive review from a novel and dedicated GNN perspective. We provide a systematic synthesis of state-of-the-art research across three fundamental dimensions: Techniques, focusing on the evolution of GNN-based methods; Applications, evaluating their deployment in diverse scenarios; and Challenges, identifying critical frontiers for future investigation. \label{fig:review}}
\end{figure*}

Recently, the development of GNNs \cite{1grl_gcn_2017,2grl_GAT_2018, 3grl_graph_sage_2017} has been particularly noteworthy and has become a powerful tool for analyzing graph-structured data. Naturally, a series of link prediction methods based on GNNs have also emerged. Despite state-of-the-art methods are rapidly evolving, corresponding analysis and summarization remain limited. While existing reviews have investigated link prediction from the distinct perspectives of classic LP methods, network specific LP methods, and graph representation learning (summarized in Fig. \ref{fig:review}). Lv et al. \cite{1survey_lv_2011}, V.Mart'mez et al. \cite{2survey_Martinez_2017}, and Laskshmi et al. \cite{3survey_Lakshmi_2018}gave an overview of exiting calssic non-GNN based link prediction techniques. From the perspective of GNNs, on the one hand, Barros et al. \cite{9survey_Barros_2021}, Dwivedi et al. \cite{8survey_Dwivedi_2023}, and Khemani et al. \cite{4survey_khemani_2024} summarized the graph representation learning  techniques, which can then be applied to link prediction as downstream task. On the other hand, some works focus on link prediction techniques within specific networks, such as, social networks \cite{7survey_alvi_2020}, dynamic graph \cite{6survey_skarding_2022}, and hyper graph \cite{5survey_chen_2024}). In summary, while previous reviews have provided valuable insights, they generally exhibit main limitations. First, comprehensive surveys focusing explicitly on the intersection of GNN and link prediction are still scarce. Second, existing graph representation learning reviews typically treat link prediction merely as a generic downstream task. This perspective often fails to capture the intricate GNN designs specifically engineered for link prediction. This study is motivated by the urgent need to bridge the gap between rapid GNN model developments and the practical requirements of complex real-world networks.

To the best of our knowledge, this is one of the first reviews to systematically explore link prediction techniques and related applications from a novel, dedicated GNN perspective. It is worth noting that we focus on models specifically designed for the link prediction problem, rather than treating it as a mere downstream task. The significance of this GNN perspective lies in its ability to trace how architectural innovations (from local message-passing to global attention) fundamentally reshape edge inference capabilities. The innovative two-dimensional taxonomy is presented in Figure 1. Specifically, we categorize the latest research into two major directions: techniques and applications, based on the inherent dichotomy between methodological innovation and scenario-driven demands. From a technique perspective, we categorize recent literature into four types based on their backbone networks: Graph Convolutional Networks (GCN), Graph Autoencoders (GAE), Graph Attention Networks (GAT), and Graph Transformers (GFormer). From an application perspective, we focus on the role of link prediction technology in the knowledge graphs and recommendation systems. Unlike existing work, we introduce a rigorous comparative framework of model strengths and limitations, making this survey a more prescriptive and actionable resource for selecting optimal GNN architectures. In addition, some open issues and future directions are also discussed. The main contributions of this paper are as follows.

\begin{itemize}
\item \textbf{Graph neural network perspective}: We examine link prediction from a novel GNN perspective, highlighting recent advancements and methods driven by GNN architectures, including GCN-based, GAE-based, GAT-based, and GFormer-based approaches.
\item \textbf{Comprehensive review}: We provide a comprehensive review of GNN-based link prediction methods from both technique and application perspectives, covering a wide variety of underlying graph types.
\item \textbf{Open issues and future directions}: We also present open issues and challenges of link prediction technologies in terms of data and methods, which offer insights for advancing future research directions.
\end{itemize}

In the rest of this survey, we give the preliminaries and notations in Section \ref{sec2}. In Section \ref{sec3}, we present a unique taxonomy for the latest research in terms of techniques and applications. Section \ref{sec4} summarizes the progress of link prediction technology based on GNNs. In Section \ref{sec5}, we explore the applications of link prediction, with a focus on the recommendation system and the KGs. Section \ref{sec6} summarizes open issues from recent research and highlights possible future directions. Finally, Section \ref{sec7} concludes this survey. 

\section{Preliminaries and notations}\label{sec2}

In this section, we provid a detailed description of preliminaries and notations. We establish a rigorous conceptual framework for GNN-based link prediction task through formalizing fundamental notation systems, systematizing graph typology classifications, paradigmaticizing link prediction task formulations, and architecturing GNNs structures. 

\subsection{Graph}
\begin{figure}[!htbp]
\centerline{\includegraphics[width=\linewidth]{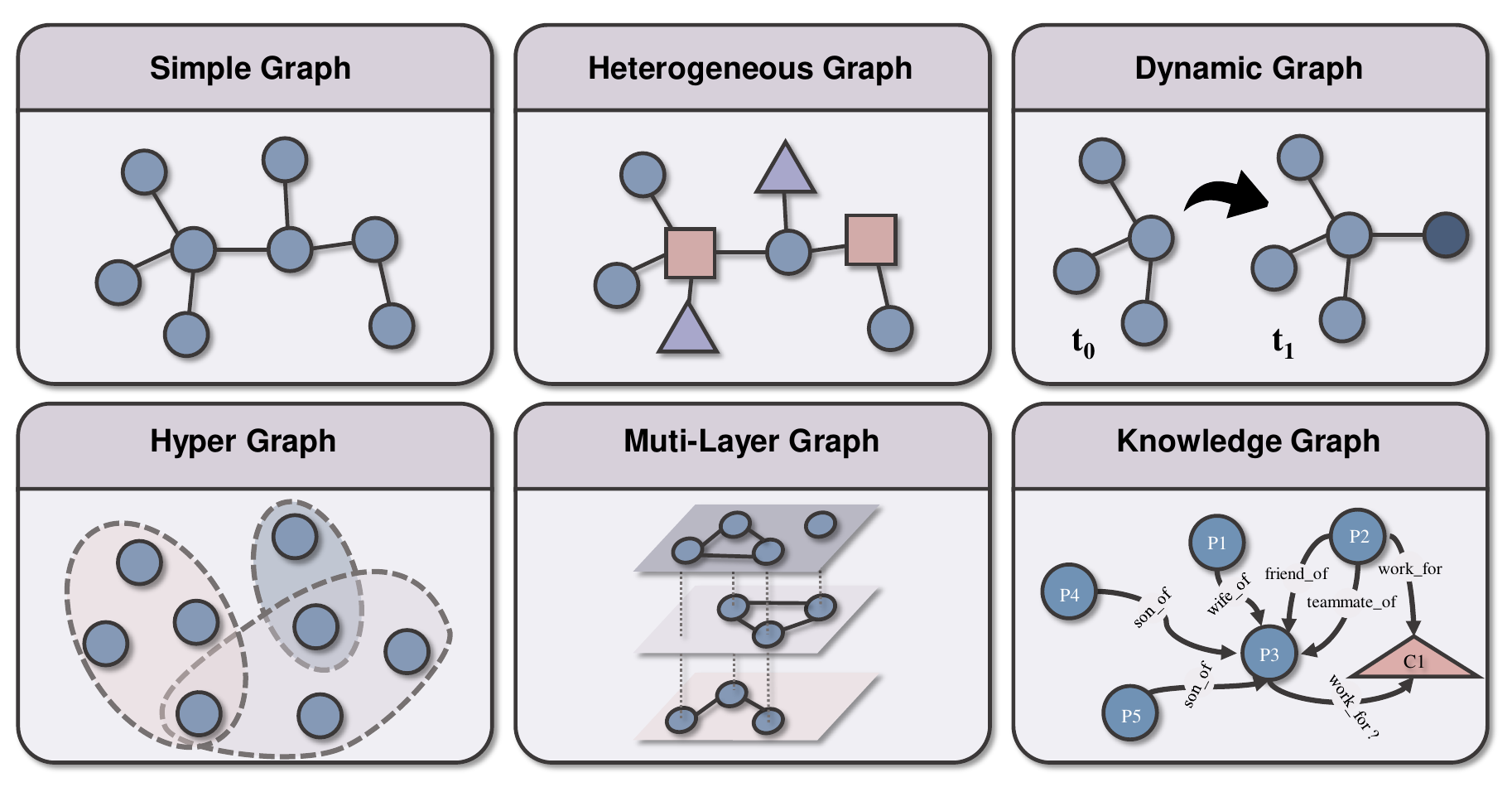}}
\caption{Taxonomy of Graph Structures in Link Prediction. This figure illustrates the diverse architectural paradigms of graphs encountered in modern link prediction, ranging from foundational to complex topologies. Simple graphs represent the basic connectivity between homogeneous nodes, while heterogeneous graphs and knowledge graphs introduce multi-type entities and relations, necessitating advanced semantic modeling. Dynamic graphs capture the temporal evolution of networks through time-stamped snapshots, whereas hypergraphs and multi-layer graphs account for high-order interactions and multi-dimensional relational strata, respectively. Together, these structures represent the multifaceted data landscape that current GNN-based methods aim to address, highlighting the transition from low-dimensional topology to high-order, non-Euclidean complexity.
\label{fig:graphs}}
\end{figure}

A graph represents objects and their relationships through a set of nodes and edges that connect them. The concept of a graph is highly flexible, enabling its application to model and analyze various types of networks and systems, including social networks, transportation networks, and computer networks. To capture the complex structure of real-world systems, research on graph-structured data has gradually expanded from simple graphs to various complex graphs. The symbolic definitions for various types of graphs are as follows, and the schematic figure is shown in Figure \ref{fig:graphs}.

\begin{itemize}
\item Simple Graph: 
A simple graph $G^{s}$ can be mathematically denoted as $G^{s} = (\mathcal{V}, \mathcal{E})$, where $\mathcal{V}$ is a set of nodes and $\mathcal{E}$ is a set of edges. Each edge is a pair of distinct nodes, i.e., $\mathcal{E} \subseteq \{\{u,v\} \mid u,v \in \mathcal{V} \text{ and } u \neq v\}$.

\item Heterogeneous Graph: 
Heterogeneous graphs allow the coexistence of different types of nodes and edges. A heterogeneous graph $G^{he}$ is represented as $G^{he} = (\mathcal{V} , \mathcal{E}, \mathcal{T}_V, \mathcal{T}_E)$, where \( \mathcal{V} \) is the set of nodes and \( \mathcal{E} \) is the set of edges. \( \mathcal{T}_V \) is the set of node types, and each node \( v \in \mathcal{V} \) has a type label \( \tau_V(v) \in \mathcal{T}_V \). Similarly, \( \mathcal{T}_E \) is the set of edge types, and each edge \( e \in \mathcal{E} \) has a type label \( \tau_E(e) \in \mathcal{T}_E \). 

\item Dynamic Graph: 
Dynamic graphs, also known as temporal graphs, contain a large amount of timestamp information.  Mathematically, a dynamic graph can be represented as: Give a graph $G^{d} = (\mathcal{V}, \mathcal{E})$, $\mathcal{V}= \{v, t_s, t_e\}$, with $v$ being a vertex of the graph and $t_s, t_e $ are respectively the start and end timestamps for the existence of the vertex (with $t_s \leq t_e$). $\mathcal{E} = \{u, v, t_s, t_e\}$, with $u, v \in \mathcal{V}$ and $t_s, t_e$ are respectively the start and end timestamps for the existence of the edge (with $t_s \leq t_e$).

\item Hyper Graph: 
Hyper Graph is a special setting of the network, which extends the concept of a graph by allowing edges to connect any number of vertices. A hyper graph \(G^{hy} \) is a pair \( G^{hy} = (\mathcal{V}, \mathcal{E}) \), where \( \mathcal{V} \) is a set of nodes and \( \mathcal{E} \) is a set of hyperedges. Each hyperedge \( e \in \mathcal{E} \) is a non-empty subset of the node set, i.e., \( e \subseteq \mathcal{V} \) and \( e \neq \emptyset \). Unlike ordinary graphs, hyperedges can connect multiple (more than two) nodes.

\item Mutilayer Graph: 
Given a set \( \mathcal{V} \) of \( n \) entities and a set \( \mathcal{L} = \{L_1, \cdots, L_\ell\} \) of layers, indexed in \( \mathcal{L} = \{1, \ldots, \ell\} \), with \( |\mathcal{L}| = \ell \geq 2 \), we denote an attributed multilayer network with \( G^{m} = (\mathcal{V}_{\mathcal{L}}, E_{\mathcal{L}}, \mathcal{V}, \mathcal{L}) \), where \( \mathcal{V}_{\mathcal{L}} \subseteq \mathcal{V} \times \mathcal{L} \) is the set of all entity occurrences, or nodes, in \( \mathcal{L} \), and in particular, \( V_l \) is the set of nodes in layer \( l \) (\( l \in \mathcal{L} \)); \( E_{\mathcal{L}} \) is the set of edges between nodes belonging to the same layer, and \( E_l \subseteq V_l \times V_l \) is the set of edges in layer \( l \). Each entity has a node in at least one layer, hence \( \mathcal{V} = \bigcup_{l=1}^\ell V_l \), and that inter-layer edges exist between each node in a layer and its counterpart in a different layer.

\item Knowledge Graph: Let \( \mathcal{V} \) denote the set of entities and \( \mathcal{E} \) denote the set of relations. A knowledge graph \( G^{k} = (\mathcal{V}, \mathcal{E}) \) represents data through a set of triplets \( (s, r, t) \), with \( s, t \in \mathcal{V} \) and \( r \in \mathcal{E} \). Consequently, a knowledge graph can be expressed as \( G^{k}(\mathcal{V}, \mathcal{E}) = \{(s, r, t) \mid s, t \in \mathcal{V}, r \in \mathcal{E}\} \subseteq \mathcal{V} \times \mathcal{E} \times \mathcal{V} \).
\end{itemize}

\subsection{Graph Neural Networks}
GNNs(Graph Neural Networks) are deep learning models designed for processing graph-structured data, effectively capturing the complex relationships and structural information between nodes within a graph. The core concept of GNNs involves updating node representations by aggregating information from their neighborhoods. The basic steps include node initialization and information aggregation. The general working principle of GNNs is given by
\begin{equation}
h_i^{(l+1)} = \sigma \left( W^{(l)} \cdot h_i^{(l)} + \sum_{j \in \mathcal{N}(i)} W^{(l)} \cdot h_j^{(l)} \right)
\end{equation}
where $h_i^{(l)}$ represents the output of node \( i \) at layer \( l \), \( \mathcal{N}(i) \) denotes the set of neighboring nodes of node \( i \), \( W^{(l)} \) is the learnable weight matrix, and \( \sigma \) is a nonlinear activation function (such as ReLU).

We analyzed GNN-based link prediction and categorized backbone networks into four main types: GCN, GAE, GAT and GFormer. We denote the adjacency matrix of a graph as \( A \), and the feature matrix of nodes is denoted as \(X\). The details are as follows.

\begin{itemize}
\item GCN

GCN \cite{1grl_gcn_2017} is a deep learning model designed for graph-structured data, utilizing an efficient layer-wise propagation rule to directly process graphs while encoding both their structure and node features. The layer-wise propagation rule is given by:
\begin{equation}
H^{(l+1)} = \sigma \left( \tilde{D}^{-\frac{1}{2}} \tilde{A} \tilde{D}^{-\frac{1}{2}} H^{(l)} W^{(l)} \right)
\end{equation}
where, \(\tilde{A} = A + I_N\) is the adjacency matrix of the undirected graph \({G}\) with added self-connections. \(I_N\) is the identity matrix, \(\tilde{D}_{ii} = \sum_j \tilde{A}_{ij}\) and \(W^{(l)}\) is a layer-specific trainable weight matrix. \(\sigma(\cdot)\) denotes an activation function, such as the \(\text{ReLU}(\cdot) = \max(0, \cdot)\). \(H^{(l)} \in \mathbb{R}^{N \times D}\) is the matrix of activations in the \(l\)th layer; \(H^{(0)} = X\). A two-layer GCN model can be defined as:
\begin{equation}
GCN(X, A) = \text{softmax}\left(\hat{A} \, \text{ReLU}\left(\hat{A} X W^{(0)}\right) W^{(1)}\right)
\end{equation}
where, $\hat{A} = \tilde{D}^{-\frac{1}{2}} \tilde{A} \tilde{D}^{-\frac{1}{2}}$ can be caculated in a pre-processing step, \(W^{(0)} \in \mathbb{R}^{D \times H}\) is an input-to-hidden weight matrix for a hidden layer with \(H\) feature maps. \(W^{(1)} \in \mathbb{R}^{H \times F}\) is a hidden-to-output weight matrix. The softmax is the activation function.

\item GAE

GAE \cite{4grl_GAE_2016} is an unsupervised learning method for graph-structured data. The GAE consists of two parts: Encoder and  Decoder. The Encoder is responsible for mapping the nodes of the graph to a low-dimensional vector space, and the Decoder's task is to reconstruct the adjacency matrix $\hat{A}$ from the embedding representation Z. The embeddings Z and the reconstructed adjacency $\hat{A}$ can be caculated by:
\begin{equation}
\hat{\mathbf{A}} = \sigma \left( \mathbf{Z} \mathbf{Z}^\top \right), \text{ with } \mathbf{Z} = \text{GCN}(\mathbf{X}, \mathbf{A})
\end{equation}

\item GAT

GAT \cite{2grl_GAT_2018} is a novel neural network architecture for graph-structured data. GAT addresses limitations of prior graph convolution-based methods by employing masked self-attention layers \cite{2024LGAT}. It enables nodes to dynamically assign varying weights to neighboring features, eliminating the need for computationally intensive matrix operations or prior graph structure information. Its core mechanism can be represented by the following formulas:
\begin{equation}
\alpha_{ij} = \frac{\exp \left( \text{LeakyReLU} \left( \mathbf{a}^T \left[ \mathbf{W} \vec{h}_i \, \| \, \mathbf{W} \vec{h}_j \right] \right) \right)}{\sum_{k \in \mathcal{N}_i} \exp \left( \text{LeakyReLU} \left( \mathbf{a}^T \left[ \mathbf{W} \vec{h}_i \, \| \, \mathbf{W} \vec{h}_k \right] \right) \right)} 
\end{equation}
\begin{equation}
\mathbf{h}'_i = \sigma \left( \sum_{j \in \mathcal{N}_i} \alpha_{ij} \mathbf{W} \vec{h}_j\right) 
\end{equation}

where, $\alpha_{ij}$ is the attention coefficient of node $j$ to node $i$, $\vec{h}_i$ and $\vec{h}_j$ are the feature vectors of nodes $i$ and $j$, $\mathbf{W}$ is the learnable weight matrix. $\mathbf{a}$ is the weight vector used for computing the attention coefficients, $\text{LeakyReLU}$ is the rectified linear unit activation function with a negative input slope. $\sigma$ is the nonlinear activation function (e.g., ReLU), $\mathcal{N}_i$ is the neighborhood set of node $i$.

\item GFormer

GFormer is a model that combines GNNs with the Transformer architecture \cite{Att-GNN2023}. The Transformer was initially proposed in the field of Natural Language Processing (NLP) \cite{5grl_GFormer_2017} for processing sequential data, relying on the self-attention mechanism to capture the relationships among elements within a sequence. Subsequently, its effectiveness for graph data has also been demonstrated. 

The core structure in Transformer is the self-attention mechanism. Let \( H = \left[ h_1, \cdots, h_n \right] \in \mathbb{R}^{n \times d} \) denote the input of self-attention module where \( d \) is the hidden dimension and \( h_i \in \mathbb{R}^{1 \times d} \) is the hidden representation at position \( i \). The input \( H \) is projected by three matrices \( W_Q \in \mathbb{R}^{d \times d_Q} \), \( W_K \in \mathbb{R}^{d \times d_K} \) and \( W_V \in \mathbb{R}^{d \times d_V} \) to the corresponding representations \( Q, K, V \). The self-attention is then calculated as:
\begin{equation}
Q = HW_Q, \quad K = HW_K, \quad V = HW_V
\end{equation}
\begin{equation}
A = \frac{Q K^\top}{\sqrt{d_K}}, \quad \text{Attn}(H) = \text{softmax}(A) V    
\end{equation}
Self-attention mechanism can be applied to GNNs in various ways, When used as an inter-layer propagation mechanism, it can be represented as:
\begin{align}
h^{'(l)} &= \text{MHA}(\text{LN}(h^{(l-1)})) + h^{(l-1)} \\
h^{(l)} &= \text{FFN}(\text{LN}(h^{'(l)})) + h^{'(l)}
\end{align}
LN refers to layer normalization, MHA denotes multi-head self-attention, and FFN represents feed-forward blocks.
\end{itemize}

\subsection{Link Prediction}

\begin{figure*}[!htbp]
\centerline{\includegraphics[width=\textwidth]{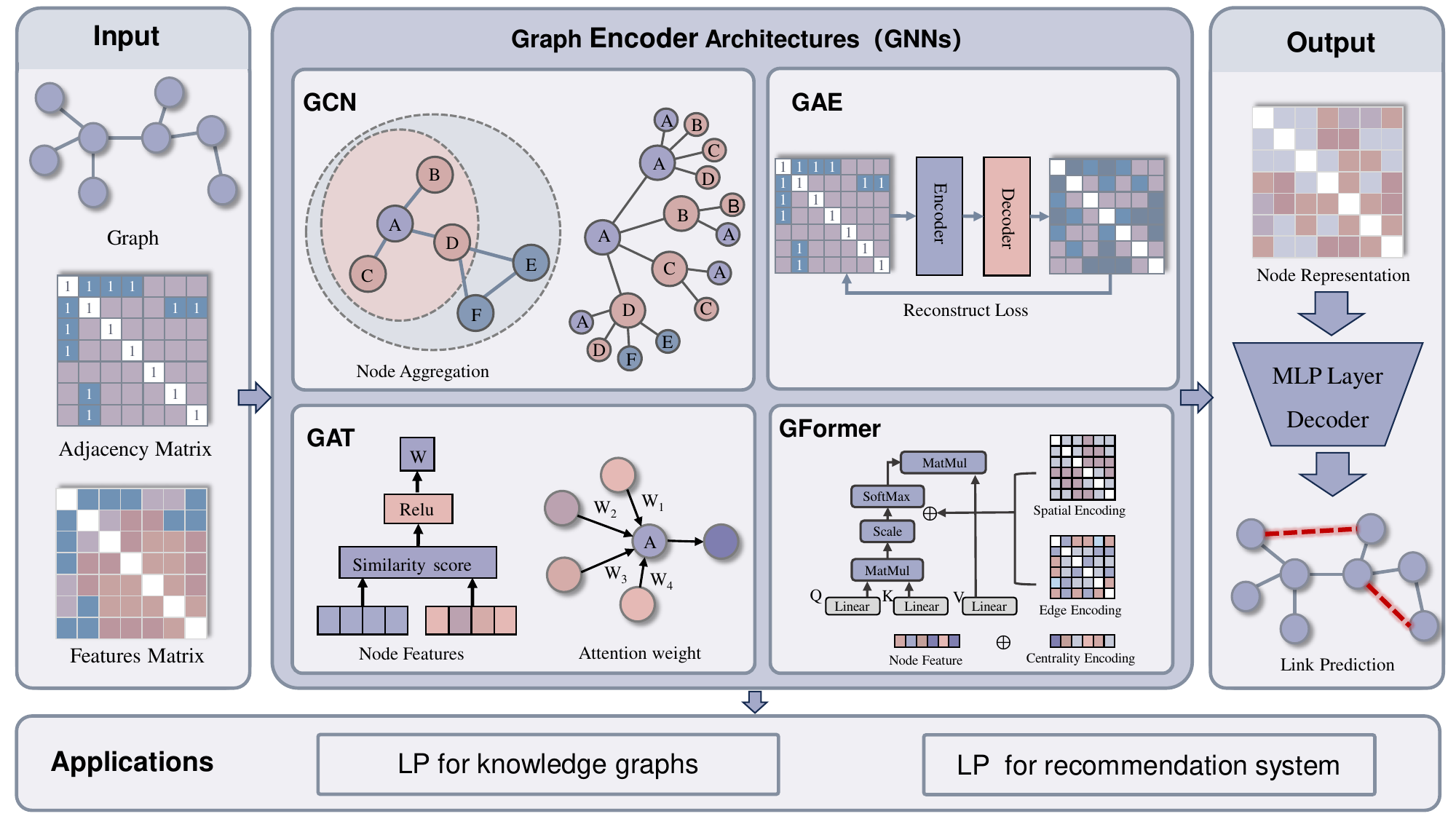}}
\caption{GNN-based link prediction framework. This figure illustrates a typical GNN-based link prediction pipeline. A GNN encoder (e.g., GCN, GAT, GAE, or GFormer) takes the adjacency matrix and node features as input to generate node representations, which are then fed into an MLP decoder to predict missing links. In the predicted graph, solid lines indicate node pairs predicted as positive (i.e., existing links), while dashed lines indicate node pairs predicted as negative (i.e., missing links).\label{fig:gnn4lp}}
\end{figure*}

Link prediction is a specialized task within the edge-level category, shown in Figure \ref{fig:gnn4lp}. Link prediction aims to predict the existing yet unknown or missing links within a graph, leveraging both the existing graph topology and node features. Let \( G^o = ( \mathcal{V}, \mathcal{E}^o ) \) represent an observed graph with \( N \) nodes, where \( \mathcal{V} \) is the set of nodes and \( \mathcal{E}^o \) is the set of observed links (edges). The observed link set \( \mathcal{E}^o \) is a subset of the set of all true links \( \mathcal{E}^* \), i.e., \( \mathcal{E}^o \subset \mathcal{E}^* \). The goal of link prediction is to infer missing links from a sample set \( \mathcal{E}^s \), which contains both true (in \( \mathcal{E}^* \)) and false (not in \( \mathcal{E}^* \)) links. Formally, design an algorithm \textsc{LinkPredictor} that takes an observed graph \( G^o \subset G \) as input and a bool value(or link probability) as output: \( \textsc{LinkPredictor}(G^o) = \Pi : \mathcal{V} \times \mathcal{V} \rightarrow \{\text{True}, \text{False}\} \) which accurately classifies links in \( \mathcal{E}^s \).

\section{Taxonomy of GNN-Based link prediction}\label{sec3}
As a fundamental task in graph analysis, link prediction has consistently attracted significant research attention. In recent years, driven by the rapid advancement of graph neural networks (GNNs), a large number of GNN-based link prediction methods have emerged. These methods leverage the powerful representation learning capabilities of GNNs and have achieved notable performance improvements. However, despite the growing body of literature, few studies have systematically reviewed these works specifically from a GNN perspective. Such a systematic review is crucial for identifying current technological trends, uncovering potential research gaps, and further advancing the field. To address this gap, this survey comprehensively summarizes recent advances in link prediction from a GNN perspective and proposes a two-dimensional fine-grained taxonomy to highlight the pivotal role of GNNs in this domain.

\begin{figure}[!htbp]
    \centering    \includegraphics[width=0.9\linewidth]{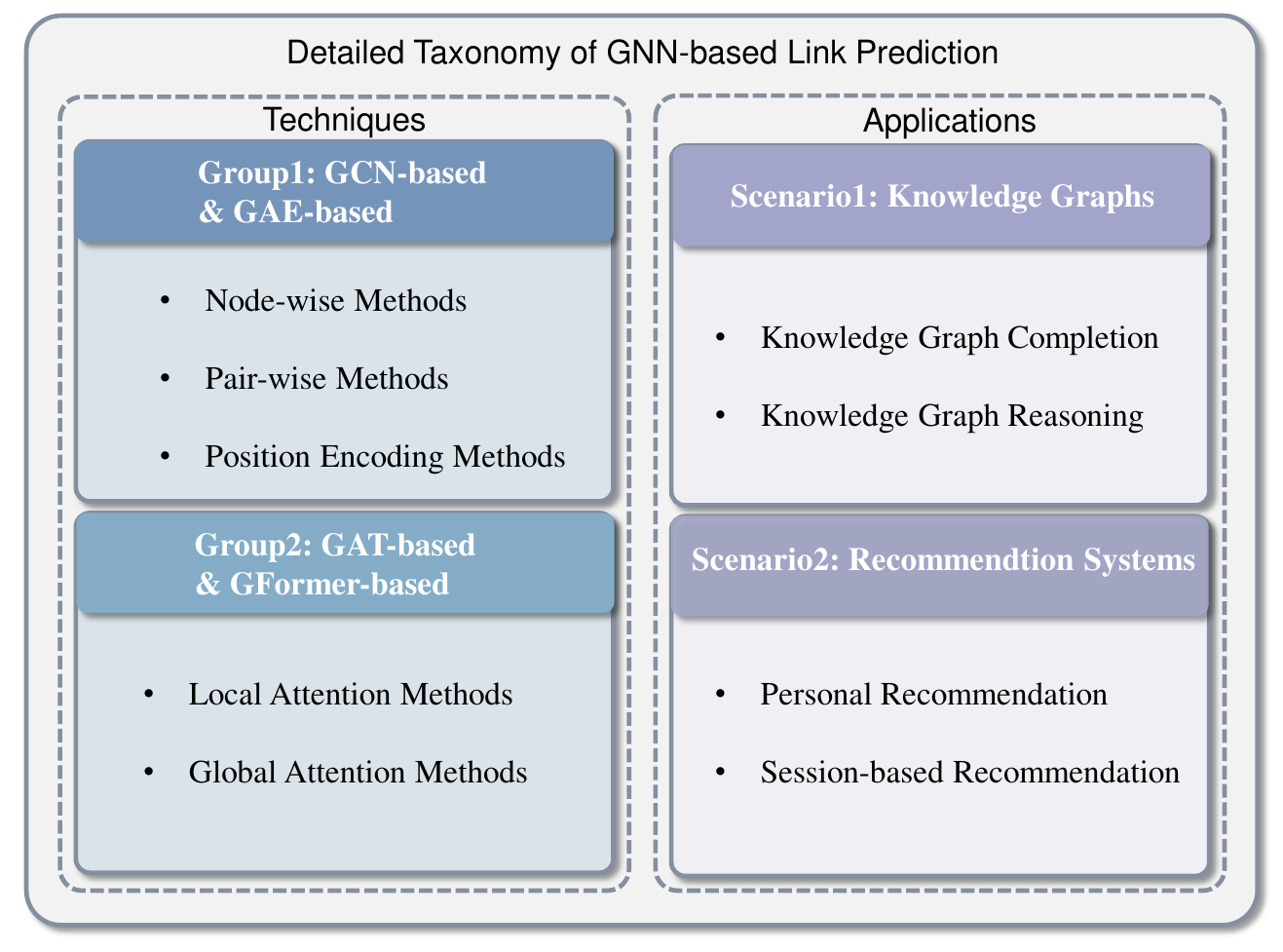}
    \caption{The detailed taxonomy of GNN-based link prediction. We categorize existing research into two major directions: techniques (left) and applications (right). Specifically, the technique direction is further subdivided into two groups: Group 1 comprises GCN-based and GAE-based methods, which focus on node-wise, pair-wise, and position encoding mechanisms; Group 2 includes GAT-based and GFormer-based methods, which emphasize local and global attention mechanisms. The application direction highlights key real-world scenarios, including knowledge graph completion and reasoning, as well as diverse recommendation systems.}
    \label{fig:taxonomy}
\end{figure}

Based on the inherent logic and objectives of this study, we categorize the existing research on link prediction into two major directions: techniques and applications, as illustrated in Figure 4. From a technique perspective (left), we focus on literature that designs and optimizes GNN models specifically for link prediction. This line of research aims to enhance the performance and efficiency of link prediction through innovative model architectures and optimization strategies. Specifically, as shown in the hierarchical structure of Figure \ref{fig:taxonomy}, the technical evolution spans a paradigm shift from basic neighborhood aggregation in GCN-based models to sophisticated adaptive attention and global dependency modeling in GFormer-based architectures. By categorizing these methods into node-wise, pair-wise, position encoding-based, and attention-based mechanisms, we highlight how different architectural designs address the intrinsic challenges of graph structural encoding. From the application perspective (right), we concentrate on the deployment of link prediction techniques in real-world scenarios. In this direction, GNN-based link prediction research demonstrates broad applicability and addresses crucial practical considerations. This detailed two-fold taxonomy ensures a comprehensive exploration of both the theoretical foundations and the empirical utility of current research. The core literature of our survey is summarized in Table 1. We provides a comparative analysis of mainstream GNN-based link prediction models, highlighting their respective strengths and inherent limitations. For example, GCN-based methods offer high efficiency and robust local aggregation but are prone to structural over-smoothing. GAE-based approaches excel in unsupervised distribution modeling, yet they often suffer from limited expressiveness and a transductive nature. To address these issues, GAT-based models introduce adaptive neighbor weighting to filter noise, while GFormer-based models leverage global context to capture long-range dependencies. However, both GATs and GFormers face significant challenges in terms of high computational costs and memory bottlenecks.

\begin{sidewaystable}
\caption{Summary of GNN-based link prediction techniques.}\label{table:GNN-based LP}
\begin{tabular*}{\textheight}{@{\extracolsep\fill}llllll}
\toprule
\textbf{Taxonomy} & \textbf{Graph Type} & \textbf{Method} & \textbf{Model} & \textbf{Year} & \textbf{Reference} \\
\midrule
\multirow{19}{*}{GCN-based LP} & \multirow{18}{*}{\(G^s\)} & \multirow{6}{*}{node-wise} & CPAGCN & 2023 &  \cite{he_community_2023} \\
                     & & & PA & 2025 &  \cite{subramonian_networked_2025}\\
                     & & & CLP& 2024 &   \cite{zhao_conformalized_2024}\\
                     & & & RGIB & 2023 &  \cite{zhou_combating_2023} \\
                     & & & TC &  2024 &  \cite{wang2024topological} \\
                     & & & LTLP & 2024 &  \cite{wang_optimizing_2024} \\
\cmidrule{3-6}
                     & &\multirow{10}{*}{pair-wise} & walkPooling & 2022 &  \cite{pan2022neural}\\
                     & & & FakeEdge & 2022 &  \cite{dong2022fakeedge} \\
                     & & & DPLP & 2024 &  \cite{ran_differentially_2024}\\
                     & & & Scaled & 2022 &  \cite{louis2022sampling} \\
                     & & &  PS2 & 2023 &  \cite{tan2023bring}\\
                     & & & LGCL & 2023 &  \cite{zhang2023line}\\
                     & & & ELPH & 2023 &  \cite{chamberlain2023graph} \\
                     & & & Proxi & 2024 &  \cite{tola2024proxi}\\
                     & & &  BloomSigLP & 2024 &  \cite{zhang_learning_2024}\\
                     & & & Eclip & 2024 &  \cite{liu_edge_2024} \\
\cmidrule{3-6}
                     & &\multirow{2}{*}{position-encoding} & HPLC & 2024 &  \cite{kim_hierarchical_2024}\\
                     & & & PEG & 2022 &  \cite{wang2022equivariant}\\
\cmidrule{2-6}
                     & \(G^{he}\)& node-wise & xGCN & 2023 &  \cite{song_xgcn_2023}\\
\cmidrule{2-6}
                     & $G^{hy}$ & node-wise & LHGNN & 2025 &  \cite{rui_higher-order_2025}\\
\midrule
\multirow{6}{*}{GAE-based LP} &  \multirow{6}{*}{\(G^s\)} & \multirow{4}{*}{node-wise} & DGAE & 2022 & \cite{wu_stabilizing_2022} \\
                     & & & CensNet-VAE &  2023 &  \cite{jiang_co-embedding_2023} \\
                     & & & D-VGAE &  2024 &  \cite{cho_decoupled_2024} \\
                     & & & Refined-GAE &  2024 &  \cite{ma_reconsidering_2024} \\
\cmidrule{3-6}
                     &  & \multirow{2}{*}{pair-wise} & Labeling Trick &  2022 &  \cite{zhang_labeling_2022} \\
                     & & & NCNC &  2024 &  \cite{wang_neural_2024} \\
\midrule
\multirow{9}{*}{GAT-based LP} & \multirow{2}{*}{\(G^{he}\)} & \multirow{2}{*}{global attention } & SIHG & 2023 &  \cite{luo2023interpretable} \\
                     & & & LPMPA & 2023 &  \cite{zhao2023link}\\
\cmidrule{2-6}
                     & \(G^m\) & local attention & ML-link & 2024 &   \cite{zangari_link_2024}\\
\cmidrule{2-6}
                     & \multirow{4}{*}{\(G^d\)} & global \& local attention & DATGN & 2023 &  \cite{mi2023double} \\
\cmidrule{3-6}
                     & & \multirow{3}{*}{local attention } & CTGLP &  2024 &  \cite{luo2024dynamic} \\
                     & & & TNCN & 2024 &  \cite{zhang_efficient_2024} \\
                     & & & IDEA & 2024 &  \cite{qin2023high} \\
\cmidrule{2-6}
                     & \multirow{2}{*}{\(G^s\)} &\multirow{2}{*}{local attention } & Disenlink & 2022 &  \cite{zhou_link_2022}\\
                     & & & Debias & 2023 &  \cite{luo2023cross}\\
\midrule
\multirow{5}{*}{GFormer-based LP} & \multirow{2}{*}{\(G^s\)}  & \multirow{2}{*}{global attention } & LPFormer & 2024 &  \cite{shomer_lpformer_2024} \\
                          &  & & SIEG & 2024 &  \cite{shi_structural_2024} \\
\cmidrule{2-6}
                          & \multirow{2}{*}{\(G^d\)} & \multirow{2}{*}{local attention } & BehaviorNet & 2024 &  \cite{liu_behaviornet_2024} \\
                          &  &  & T-SPEAR & 2024 &  \cite{lee2024spear} \\
\cmidrule{2-6}
                          & \(G^{he}\) & global attention  & Path2pair & 2024 &  \cite{hang_paths2pair_2024} \\
\botrule
\end{tabular*}
\end{sidewaystable}

\textbf{Technique perspective}: GNNs have proven effective in link prediction, leading to the emergence of various methods. These methods mainly use a message-passing mechanism and can be categorized into four types based on their backbone networks: GCN-based, GAE-based, GAT-based, GFormer-based. GCN updates node embeddings by aggregating information from their neighbors, but its limitation lies in its strong assumption of graph homophily. Therefore, GCN-based methods are mostly used for homogeneous graphs. GAE demonstrates stronger applicability and efficiency in unsupervised learning and large-scale graph data processing. It is worth noting that, in this survey, GAE-based methods refer to those that employ GAE as the backbone model. GAT introduces an attention mechanism that allows the model to dynamically assign different weights to different neighbors. Thus, GAT-based methods are well suited for graphs with distinct node features or complex structures. Owing to its effectiveness, the Transformer architecture has been increasingly adopted for graph data processing and widely applied to link prediction tasks. GFormer-based methods encompass approaches that integrate Transformer technology into GNNs. 

\textbf{Application perspective}: Given their representative significance and practical value, KGs and recommendation systems receive significant attention in this survey. KGs utilize link prediction to complete missing relationships or infer potential connections, widely applied in fields such as semantic search and intelligent question answering. Recommendation systems, on the other hand, enhance the accuracy of personalized recommendations by predicting potential interactions between users and items. These two domains not only demonstrate substantial practical significance but also present critical technical challenges, including data sparsity and dynamic behavior modeling, which have garnered considerable attention in cutting-edge research. Particularly in recent years, with the advancement of GNNs, the application of link prediction in KGs and recommendation systems has seen remarkable progress, becoming a focal point of research. 

\section{Technique perspective}\label{sec4}

\subsection{GCN-based Link Prediction}
GCN is an efficient graph learning model commonly used for link prediction, particularly in static homogeneous graphs.  GNN-based link prediction methods employ message-passing mechanisms to iteratively aggregate neighborhood information, integrating both node attributes and topology information to learn effective latent node embeddings. Following this paradigm, recent research have enhanced the expressive power of GCN specifically tailored for link prediction scenarios in three ways: node-wise approaches \cite{he_community_2023,subramonian_networked_2025,zhao_conformalized_2024, song_xgcn_2023, zhou_combating_2023, wang2024topological, rui_higher-order_2025}, pair-wise approaches \cite{zhang_learning_2024,liu_edge_2024,ran_differentially_2024,pan2022neural,chamberlain2023graph, dong2022fakeedge, tola2024proxi, tan2023bring, louis2022sampling, zhang2023line} and position-encoding approaches \cite{kim_hierarchical_2024,wang2022equivariant}. Node-wise methods focus on learning representations for individual nodes and then combining the representations of node pairs for prediction, emphasizing the feature information of nodes and their neighbors. In contrast, pair-wise methods directly focus on the complex structural information between node pairs, such as the subgraphs structures centered around node pairs or common neighbors. These structural features are crucial for link prediction. Position-encoding methods capture the position information of nodes in a graph to distinguish isomorphic nodes.

As a significant instantiation of GCN-based link prediction, node-wise approaches remain prevalent in contemporary research.  Node-wise link prediction methods leverage GCN's node-centric characteristics to learn representations for the two end nodes of a link, obtaining link representations through concatenation or summation. CPAGCN \cite{he_community_2023} found that nodes within the same community or with greater community overlap are more likely to form links. 
Therefore, the author integrates the node representation matrix obtained through GCN with an overlapping community detection model to reconstruct the graph. 
This method improves link prediction accuracy by leveraging community-shared information via loss of graph construct. Subramonian et al. \cite{subramonian_networked_2025} first discovered that GCNs using a symmetric normalized graph filter exhibit a preferential attachment (PA) bias within the group, a tendency in link prediction tasks to link nodes with high degrees. To tackle the group-level unfairness resulting from this PA bias, the article introduces a fairness regularization term into the original GCN's training loss function. By refining the structure of GNNs and introducing a hybrid aggregator, Rui et al. \cite{rui_higher-order_2025} achieved enhanced performance in higher-order link prediction while also streamlining the model architecture. To construct a reliable prediction interval for GCN-Based link prediction, Zhao et al.  \cite{zhao_conformalized_2024} incorporated conformal prediction techniques into link prediction. Additionally, sampling methods were employed to adjust the degree distribution of the graph nodes to approximate a power-law distribution, thereby enhancing the efficiency of conformal prediction. Wang et al. \cite{wang2024topological} assign weights to edges based on the local subgraph differences between nodes, enabling nodes to obtain information from nodes with stronger topological interactions. RGIB \cite{zhou_combating_2023} enhances model robustness against noise by balancing the mutual information between input graph topology, target labels, and learned representations. Wang et al. \cite{wang_optimizing_2024} proposed a data augmentation method that can be used for both node-wise and pair-wise methods. LTLP \cite{wang_optimizing_2024} introduced the long-tail problem in link prediction: pairs with many common neighbors are a minority. A subgraph enhancement module is designed to improve link prediction performance by increasing the number of common neighbors of the tail nodes. Unlike other methods, xGCN  \cite{song_xgcn_2023} utilizes an unsupervised link prediction approach for heterogeneous graphs, enhancing the efficiency and scalability of the GCN.

\begin{figure}[!htbp]
\centerline{\includegraphics[width=\linewidth]{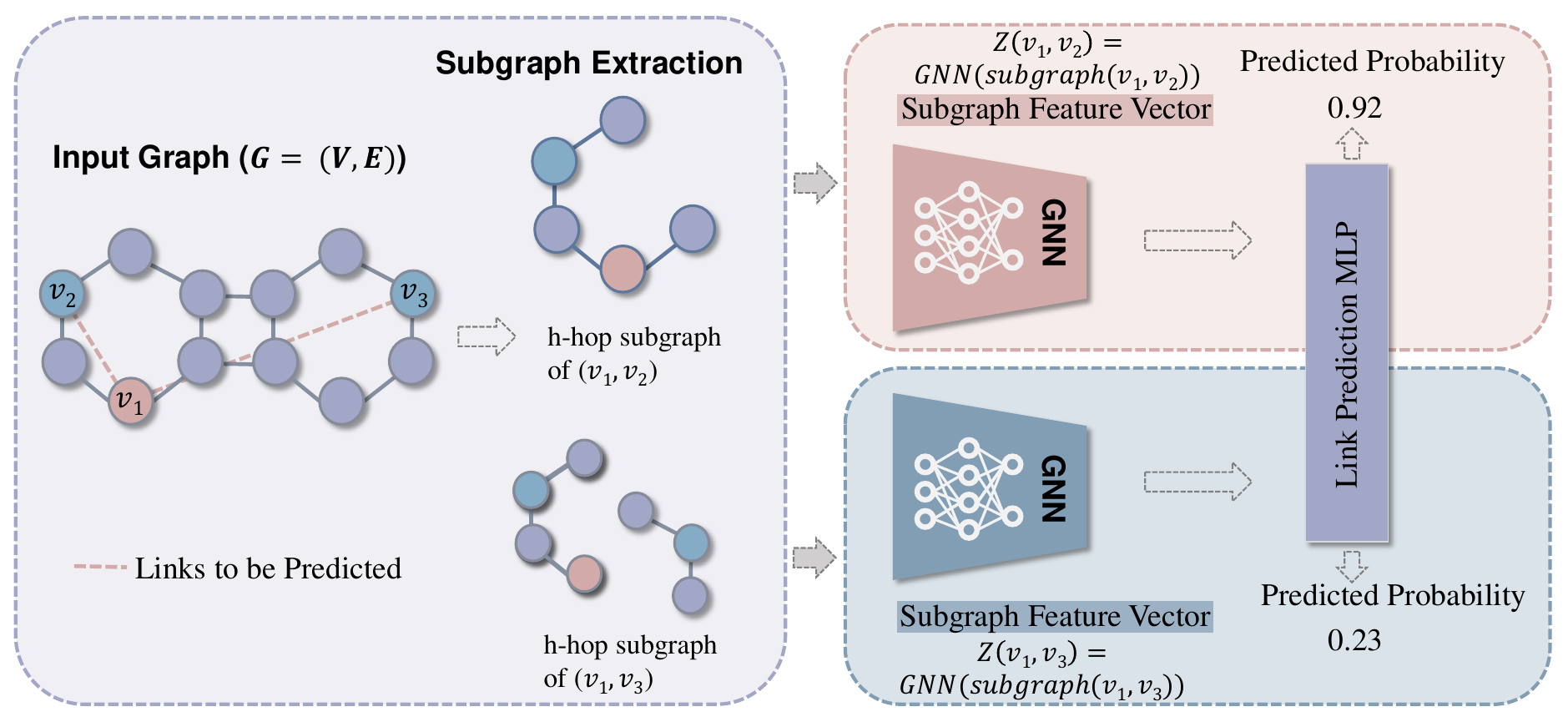}}
\caption{Coupled pair-wise method. Main coupled pair-wise methods is subgraph-based, leveraging GNNs to learn the subgraph structure around links.\label{fig:coupled}}
\end{figure}

\begin{figure}[!htbp]
\centerline{\includegraphics[width=\linewidth]{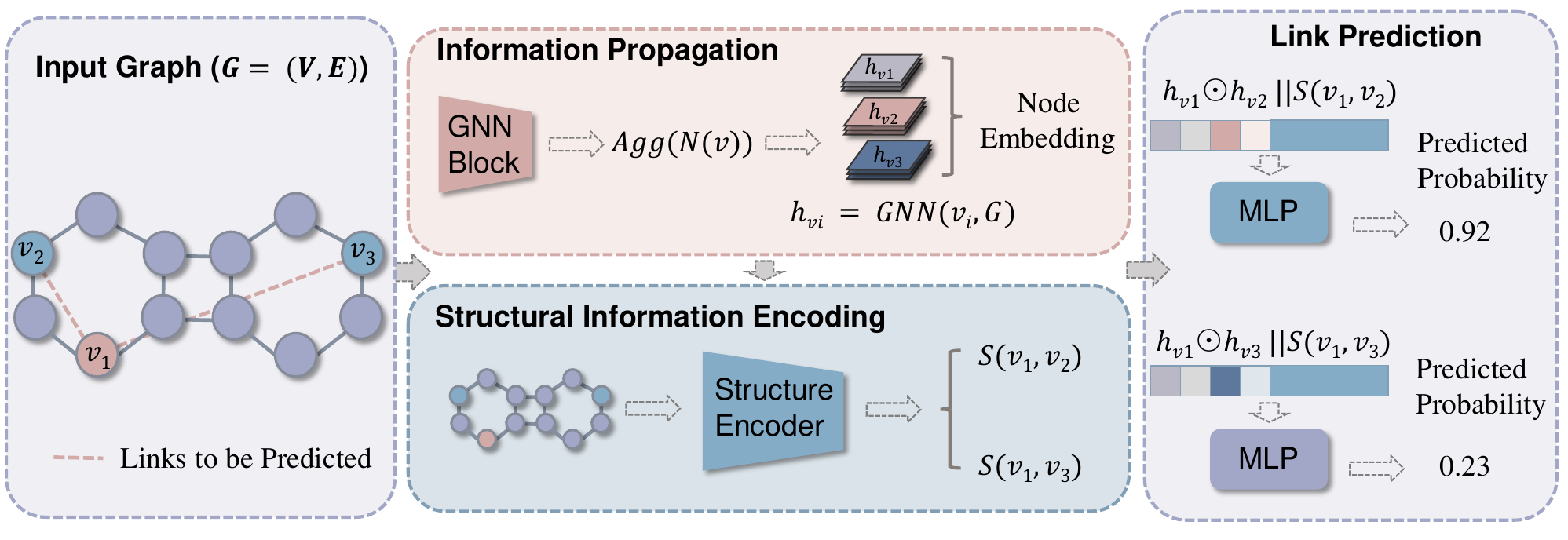}}
\caption{Decoupled pair-wise method. Most decoupled pairwise methods build upon node-wise methods by adding pairwise structural information. The design focus of these methods lies in the pairwise feature encoder.\label{fig:decoupled}}
\end{figure}

Pair-wise methods represent another important approach to GCN-based link prediction, which have garnered substantial attention for their ability to directly model the intricate structural relationships between node pairs. GCN sometimes shows suboptimal results in link prediction, which can be largely attributed to its neglect of information specific to target links. For example, it may not learn the similarity between node pairs like link heuristics do, Chen et al.  \cite{Chen_count_2020} proves that GNNs are unable to count connected substructures such as triangles. To enhance the expressive ability of vanilla GCN, many methods have incorporated link-specific structural information into GCN. Pair-wise methods can be further divided into coupled and uncoupled methods, the difference between these two categories can be seen in Figure \ref{fig:coupled} and Figure \ref{fig:decoupled}. A common and effective coupled approach is to extract the local subgraph of the target link and classify it using GCN \cite{pan2022neural,dong2022fakeedge,ran_differentially_2024,louis2022sampling,tan2023bring}. However, they extract the subgraph in different ways. WalkPooling  \cite{pan2022neural} and FakeEdge \cite{dong2022fakeedge} captures rich structural information by sampling k-hop neighborhood subgraphs. To reduce data dependency, Ran et al. \cite{ran_differentially_2024} propose an innovative path subgraph extraction method to replace the neighborhood subgraph. ScaLed  \cite{louis2022sampling} employs random walks for subgraph sampling. Furthermore, Tan et al. \cite{tan2023bring} innovatively parameterize the subgraph selection process, enabling it to automatically learn and infer the optimal subgraph structure based on the features of the edges. As an innovative edge representation method, subgraph extraction has also been introduced into a contrastive learning framework \cite{zhang2023line}. Zhang et al. \cite{zhang2023line} improves performance in link prediction tasks by maximizing the mutual information between edge representations obtained through subgraphs and line graphs. Subgraph extraction-based methods have achieved significant performance. However, explicitly constructing subgraphs is computationally expensive. So, many researchers try to design decoupled models. To reduce computational burden, these decoupled methods encode structural features and integrate them as additional features to enhance GCN \cite{chamberlain2023graph,tola2024proxi,zhang_learning_2024}. BloomSigLP \cite{zhang_learning_2024} encodes node neighborhoods by compressing neighborhood information into compact bit arrays using hashing techniques, and recovers structural features such as neighborhood overlap through simple bitwise operations. ELPH \cite{chamberlain2023graph} uses subgraph sketches to estimate the intersection and union of node neighborhoods, allowing the rapid computation of structural features during training and inference. PROXI  \cite{tola2024proxi} encodes structural indices and domain indices into a set of features and combines them with the node pair representations learned by GCN to predict the existence of links.  However, the aforementioned methods all overlook the rich information contained in the edges of the graph. To fill this gap, ECLip \cite{liu_edge_2024} integrates edge information into node embeddings using a contrastive learning framework and feeds the node embeddings into a pair-wise predictor for link prediction. 

Position-encoding approaches is a technique for mapping nodes in a graph to a low-dimensional vector space, where these vectors capture the position information of the nodes within the graph. Position-encoding improves the GNN understanding of graph structure and differentiates isomorphic nodes. HPLC  \cite{kim_hierarchical_2024} integrate landmark selection with graph clustering to assign location information to nodes to distinguish various links. PEG  \cite{wang2022equivariant} generates permutation-equivariant positional encodings by separately updating node and position features. 

\subsection{GAE-based Link Prediction}

GAE is an advanced encoder-decoder GNN designed for efficient processing of graph data. By minimizing reconstruction loss, GAE learns effective representations, ensuring scalability for large datasets. The node embeddings produced by GAE aid in predicting unobserved links, making it popular for link prediction tasks. Similar to GCN, GAE-based methods are mostly used for homogeneous graphs. 

Recently, many researchers have made improvements to GAE for the link prediction task. Ma et al. \cite{ma_reconsidering_2024} optimized GAE through hyperparameter tuning and techniques like orthogonal embedding and linear propagation, achieving the performance of more complex models while improving computational efficiency. A hybrid structural model for link prediction was designed by combining CensNet and VAE in CensNet-VAE \cite{jiang_co-embedding_2023}. The structure employs CensNet in the encoder, which is a generic graph embedding architecture consisting of node and edge layers. D-VGAE \cite{cho_decoupled_2024} highlights the limitations of the homogeneity assumption in GAE for link prediction tasks. It proposes a hard Expectation-Maximization (EM) algorithm for end-to-end learning by integrating two distinct embedding spaces: a homogeneity space based on cosine similarity and a node popularity space based on the norm. Most GAE-based link prediction techniques rely on its shallow structure. DGAE \cite{wu_stabilizing_2022} breaks through the limitations of traditional GAE-based link prediction methods. It is the first to enhance the performance of GAE by deepening its structure, thereby more effectively extracting information from higher-order neighborhoods and node features. By constructing residual connections and using autoencoders for co-embeddings, the model captures multi-scale information and generates a more compact joint edge-node representation, significantly improving the accuracy and robustness of link prediction.

These methods have made improvements to GAE and have boosted the performance of link prediction in various ways. However, their fundamental approach is still to first apply GAE to the entire network to calculate the feature representation of each node and then aggregate the features of the two end nodes of a link to predict the likelihood of the link. However, this approach prevents GAE from capturing the interdependencies between the two nodes at the ends of a link, and it fails to distinguish links with different structural roles within the graph. Labeling Trick \cite{zhang_labeling_2022} and NCNC \cite{wang_neural_2024} have explored the causes of this limitation and have enhanced GAE's ability to represent multiple nodes effectively. Zhang et al. \cite{zhang_labeling_2022} leverage target-nodes-distinguishing and permutation equivariance properties of the labeling trick to learn the most expressive structural representation for a set of nodes. This enables GAE to assign the same representation to isomorphic links while differentiating all non-isomorphic links. Wang et al. \cite{wang_neural_2024} guide the pooling step of the original GAE-based link prediction method by combining the node representations of the target link's end nodes and their common neighbors. This approach helps GAE distinguish between non-isomorphic links formed by isomorphic nodes.

\subsection{GAT-based Link Prediction}

\begin{figure}[!htbp]
\centerline{\includegraphics[width=\linewidth]{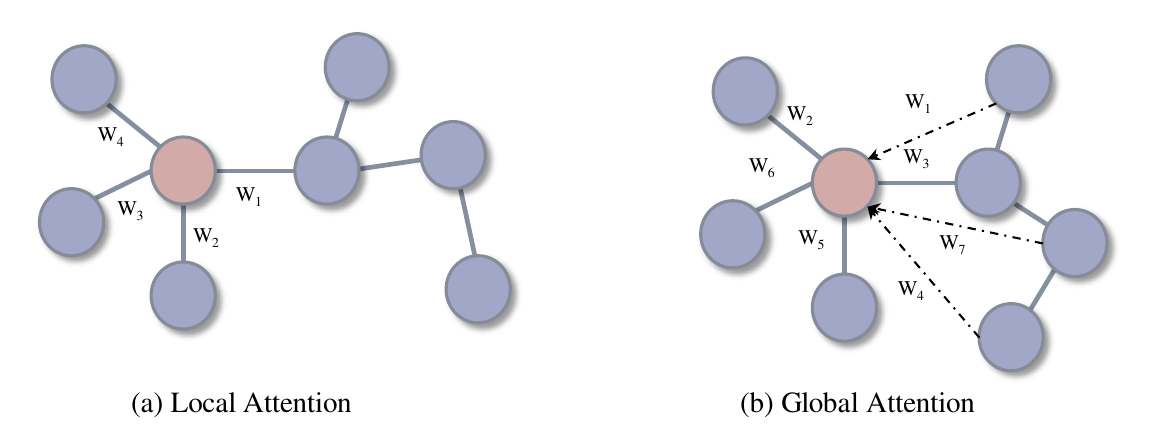}}
\caption{Two types of attention mechanisms in GNN-based link prediction methods. Local attention focuses only on the direct neighbors of the target node, which global attention further involves higher-order neighbors.\label{fig:attention}}
\end{figure}

GCN and GAE typically assume that all neighboring nodes contribute equally when aggregating neighborhood information. This approach is straightforward and efficient. However, these models face challenges when dealing with complex and heterogeneous structures. In contrast, GAT employs an attention mechanism that allows it to dynamically assign different weights to neighboring nodes based on their importance, as shown in Figure \ref{fig:attention}. This enables the model to selectively focus on more relevant neighbors when aggregating information, thereby enhancing its ability to capture complex relationships within the graph. 

Given that GAT can distinguish between different types of nodes and edges, numerous GAT-based techniques have recently been widely applied to heterogeneous graphs \cite{zangari_link_2024,zhao2023link,luo2023interpretable}. ML-Link \cite{zangari_link_2024} employs the attention mechanism in multilayer heterogeneous graphs to dynamically assign weights to different types of overlapping neighbor, highlighting the significance of different neighborhoods more effectively. SIHG \cite{luo2023interpretable} successfully employs GAT for signed heterogeneous graphs by integrating a sign attention module. This approach maximizes mutual information between edge polarities and node embeddings, facilitating the recognition of the most representative neighboring nodes for edge sign inference. Zhao et al. \cite{zhao2023link} employ a local attention mechanism to dynamically aggregate the embeddings of the node pair of different metapaths, effectively capturing the significance of local semantic information in heterogeneous networks to improve the accuracy of link prediction.

Additionally, GAT has also demonstrated remarkable performance in dynamic graphs \cite{zhang_efficient_2024,mi2023double,luo2024dynamic,qin2023high}.
CTGLP \cite{luo2024dynamic} incorporates a time-aware local attention mechanism that assigns higher aggregation weights to neighboring nodes with more recent interactions with the central node. This method effectively captures temporal dependencies in dynamic graphs, significantly enhancing the performance of dynamic link prediction. DATGN \cite{mi2023double} introduces global and local dual-attention mechanisms to capture the global temporal information and local spatiotemporal evolution patterns in dynamic graphs, demonstrating the unique value of attention mechanisms in dynamic graph analysis. Qin et al. \cite{qin2023high} introduce an attention mechanism to dynamically align node hidden states across different time steps in dynamic graphs, effectively handling changes in node sets and enhancing the model's adaptability and prediction accuracy. TNCN \cite{zhang_efficient_2024} also successfully applies graph attention to link prediction in dynamic graphs. This method enhances the performance of dynamic link prediction by efficiently extracting temporal neighboring structures and obtaining multi-hop common neighbor information. Notably, the method employs graph attention embeddings to process node memory, thereby acquiring temporal representations. 

Most significantly, GAT can also effectively address more complex issues within homogeneous graphs \cite{zhou_link_2022,luo2023cross}. For instance, zhou et al. \cite{zhou_link_2022} creatively introduces the disentangled representation learning for link prediction in heterophilic graphs. Typically, homogeneous graphs generally conform to the homophily hypothesis, which suggests that nodes exhibiting similar attributes are more likely to be connected. However, this assumption does not always hold in real-world networks. Heterogeneity refers to the tendency of nodes to connect to nodes with different characteristics or class labels. Moreover, luo et al. \cite{luo2023cross} propose a twin-structure framework to alleviate the bias between cross-links and internal-links through data augmentation and embedding fusion. The framework uses GAT as its GNN model, leveraging its dynamic attention mechanism to generate effective node embeddings and improve the capture of inter-community connections. 

\subsection{GFormer-based Link Prediction}
In recent years, Transformer \cite{Att-GNN2023} architecture has achieved remarkable results in various fields. Graphormer \cite{ying_transformers_2021} shows that the Transformer architecture is not only a dominant choice in natural language processing \cite{BERT_2019,Liu_2019_RoBERTaAR} and computer vision \cite{VIT_2021,Liu_2021_SwinTransformer}, but also attain excellent results on graph representation learning. Compared to GAT, the self-attention of the transformer allows it to simultaneously consider relationships between all nodes in the graph, rather than just those within local neighborhoods. This global perception enables the Transformer to capture long-range dependencies between nodes, which is crucial for link prediction since future links may be influenced by nodes that are far apart. The unique strengths of Graph Transformer facilitate its extensive application in a variety of complex graphs \cite{shomer_lpformer_2024,shi_structural_2024,hang_paths2pair_2024,liu_behaviornet_2024,lee2024spear}. 

LPFormer \cite{shomer_lpformer_2024} and SIEG \cite{shi_structural_2024} utilize transformer to encode the structural relationships between target node pairs, successfully applying the transformer architecture to homogeneous graphs. LPFormer \cite{shomer_lpformer_2024} employs an attention module to capture the diverse structural features that contribute to link formation, alleviating the inductive bias caused by previously classifying all links using the same fundamental factors. Similarly, SIEG \cite{shi_structural_2024} introduces the Binary Structural Transformer to learn different heuristics for target links. In heterogeneous graphs, transformers can also play a significant role. Paths2Pair \cite{hang_paths2pair_2024} introduces two important modules, Meta-Path Level Aggregation and Pair Level Aggregation, in the heterogeneous graph link prediction framework. The former utilizes transformer structure to aggregate the information of each type of meta-path to the target node pairs, while the latter aggregates the interactions between different meta-paths. This approach effectively reduces the complexity and information loss challenges often encountered by traditional GNNs when handling heterogeneous graphs. BehaviorNet \cite{liu_behaviornet_2024} and T-SPEAR \cite{lee2024spear} innovatively applies the transformer architecture to dynamic graphs. By leveraging the self-attention mechanism of the Transformer, liu et al. \cite{liu_behaviornet_2024} incorporate edge behaviors as supplementary information into node representations, thereby enhancing the structural information of nodes. Additionally, the integration of fine-grained temporal information as auxiliary attributes into the graph representation contributes to improved performance in dynamic graph link prediction. T-SPEAR \cite{lee2024spear} is an adversarial attack method for link prediction on continuous-time dynamic graphs. It employs a Transformer-enhanced surrogate model to estimate edge formation probabilities and injects highly improbable edges as adversarial examples, subtly disrupting the target model's predictions.
\section{Application perspective}\label{sec5}

In the real world, many systems can be abstracted as graphs, and many problems within these systems can be modeled as link prediction tasks. Efficient link prediction not only deepens our understanding of network construction and evolution \cite{gou_triad_2022}, but also unlocks broad applications across multiple fields. In KGs, link prediction helps complete missing or new relations \cite{zhang_disconnected_2023, liu_learning_2023}, enhancing our understanding of their dynamics. Moreover, it also enhances the effectiveness of KGs in intelligent query answering scenarios \cite{arakelyan_adapting_2023}. In social networks, link prediction technology is widely applied in friend recommendations  \cite{song_friend_2022, santos_link_2021}, personalized recommendations \cite{yang_hagerec_2020,ma_temporal_2024, mao_ultragcn_2021}, and session-based recommendations (SBR) \cite{zhang_beyond_2024,zhang_bi-preference_2024,xia_efficient_2023}. By predicting possible connections between users, these systems can provide more accurate social relationship recommendations and personalized content delivery, greatly enhancing the user experience. In biological networks, identifying interactions between entities can be expensive and time-consuming. Link prediction technology can lower experimental costs and accelerate the understanding of these networks through accurate predictions \cite{clauset_hierarchical_2008,redner_teasing_2008}. Link prediction can predict drug-drug interactions \cite{long_pre_2022}, supporting combination therapies and accelerating drug development. This section focuses on analyzing the recent applications of link prediction technology in KGs and recommendation systems, highlighting innovative practices and their value.

\subsection{Link prediction for knowledge graph}

KGs are networks representing real-world entities and their relationships. As a fundamental infrastructure for structured knowledge representation, KGs effectively model real-world semantic relationships through triple structures (head entity-relation-tail entity). Their robust semantic reasoning capabilities, demonstrated in tasks such as semantic parsing  \cite{berant-2013-semantic,heck-2013-leveraging}, scene graph generation  \cite{zareian-2020-bridging}, and intelligent question-answering(QA) systems  \cite{Huang-2019-Knowledge,arakelyan_adapting_2023}, have significantly advanced cognitive intelligence. However, KGs often suffer from incompleteness due to three primary factors: (1) the dynamic evolution of entity relationships hindering real-time updates, (2) information loss during the integration of multi-source heterogeneous data, and (3) technical limitations in explicitly capturing implicit relationships during knowledge acquisition. Such structural deficiencies severely impair KG performance in downstream applications. For instance, missing critical relational chains may lead to erroneous reasoning outputs in QA systems.

\begin{figure*}[!htbp]
\centerline{\includegraphics[width=0.8\linewidth]{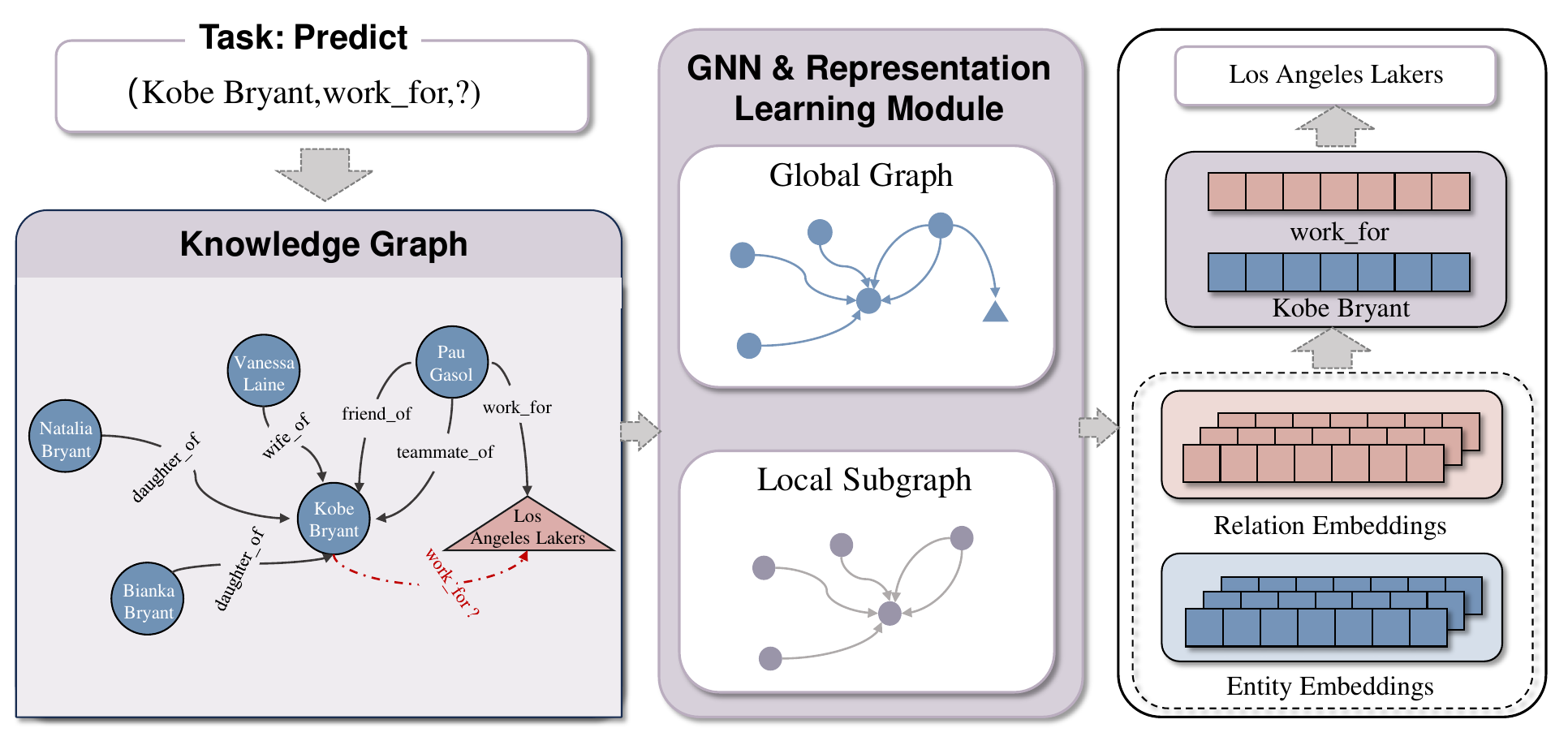}}
\caption{GNN-based link prediction framework for knowledge graph (KG) completion. The pipeline translates a query triple (h,r,?) into a downstream inference task by leveraging multi-relational context. Specifically, a GNN-based encoding module processes the KG to simultaneously capture global topological patterns and local subgraph structures, generating high-dimensional latent embeddings. By projecting entities (e.g., Kobe Bryant) and relations (e.g., work for) into a unified vector space, the framework computes the likelihood of candidate tail entities, thereby completing the missing link (e.g., Los Angeles Lakers) based on the learned structural representations.}
\label{fig:GNN4KGC}
\end{figure*}

Knowledge Graph Completion (KGC) addresses these gaps by predicting missing triples to enhance semantic networks. It strengthens the logical completeness of KGs to support complex reasoning tasks. As the core technical of KGC, link prediction models latent interaction patterns between entities to effectively infer unobserved relationships. The performance of link prediction directly determines both the accuracy and scalability of knowledge completion. Recent years have witnessed remarkable breakthroughs in KG link prediction techniques, with advanced methodologies substantially improving prediction reliability and computational efficiency. We illustrate the GNN-based KGC pipeline in Fig. \ref{fig:GNN4KGC}.

Link prediction has always been a crucial task in KGs. Inductive link prediction has gained significant attention in recent years because of its practicality and challenges \cite{li_causal_2024,zhang_disconnected_2023,zhang_inductive_2024,liu_learning_2023,mohamed_locality-aware_2023}. Inductive link prediction is a key reasoning task in KGs, aiming to predict relationships between entities that do not appear in the training data, rather than relying solely on known entities and relationships. This task is especially important for dynamically evolving KGs. FAGA \cite{li_causal_2024}, LCILP \cite{mohamed_locality-aware_2023} and REST \cite{liu_learning_2023} achieve inductive learning by extracting subgraphs around target triples and using a GNN to score the subgraphs. Unlike previous subgraph extraction methods, LCILP \cite{mohamed_locality-aware_2023} modifies the subgraph sampling approach by employing a locality-aware clustering method based on Personalized PageRank to sample densely related subgraphs around target links, thereby improving prediction efficiency. Additionally, FAGA \cite{li_causal_2024} identifies the causal relationship between subgraph semantic patterns and relation prediction, overcoming the limitations of redundant information. REST \cite{liu_learning_2023} adopts a single-source initialization method, initializing edge features only for target links, which ensures the relevance of the mined rules to the target links. In contrast, DEKG-ILP \cite{zhang_disconnected_2023} and ISE2 \cite{zhang_inductive_2024} focus on the characteristics of emerging KGs to enhance the performance of inductive link prediction. DEKG-ILP \cite{zhang_disconnected_2023} optimizes closed link and bridging link through two modules, CLRM and GSM, significantly improving the performance of inductive link prediction for discrete emerging KGs. ISE2 \cite{zhang_inductive_2024} effectively captures the long-term dependency interactions and incremental features between emerging KGs at different stages by incorporating two modules, SEFM and AGQ, which take into account the sequential-emerging nature of KGs. In addition to inductive learning, some special scenario settings have also been explored. For instance, CEKFA \cite{wang_canonicalization-enhanced_2023} extends the traditional normalization of noun phrases to the normalization of relation phrases and triples. Compared to the traditional method of decomposing n-ary relational facts into triples, NaLP \cite{guan_link_2019} models n-ary data by representing each n-ary relational fact as a set of role-value pairs, explicitly capturing the correlations among them. This approach avoids the information loss associated with traditional decomposition into triples. KGs often contain some noisy information, NYLON \cite{yu_robust_2024} innovatively introduces element-wise and fact-wise confidences via a “least confidence” principle, and combines it with active crowd learning, to investigate knowledge graph link prediction technology in noisy environments for the first time. Another technique for link prediction in KGs is optimizing knowledge graph representation learning \cite{luo_dhge_2023,wang_enhance_2023,deng_knowledge_2024,lu_schema-aware_2024,chung_learning_2023}. There are many sub-tasks of link prediction on KGs. In order to unify the input forms, contextual modeling of specific tasks, and encoding of topological information, UniLP \cite{liu_unilp_2024} utilizes generative pre-trained language models to provide a unified framework for different link prediction sub-tasks.

Beyond predicting relationships between entities, link prediction also plays a vital role in knowledge question answering on incomplete KGs \cite{arakelyan_adapting_2023,minervini_complex_2022}. CQD \cite{minervini_complex_2022} breaks down complex queries into differentiable optimization problems and uses a pre-trained neural link predictor to calculate the truth value of each atomic query, achieving efficient answers to complex queries. Subsequently, Arakelyan et al. \cite{arakelyan_adapting_2023} effectively improved the accuracy of answering complex queries on incomplete knowledge graphs by recalibrating neural link prediction scores. These studies have significantly enhanced the capacity for knowledge question answering on incomplete KGs.

\subsection{Link prediction for recommend system}

Recommendation systems are essential technologies for modern internet platforms, significantly improving user experience and platform value by predicting users' interests in products or content. However, with the increasing complexity and dynamism of user behavior data, traditional recommendation methods face numerous challenges, such as data sparsity, cold start issues, and capturing dynamic preferences. In recent years, GNNs and their applications in link prediction have provided new ideas for addressing these issues. In the context of recommendation systems, link prediction can be transformed into predicting potential interaction relationships between users and items, such as purchases, views, or clicks. GNN-based link prediction methods can effectively capture the complex relationships between users and items by learning node embeddings in the graph structure while handling heterogeneous information and dynamic behaviors. In this section, we will explore the application of GNN-based link prediction in recommendation systems, focusing on GNN-based recommendation methods. 

\begin{figure*}[!htbp]
\centerline{\includegraphics[width=0.8\linewidth]{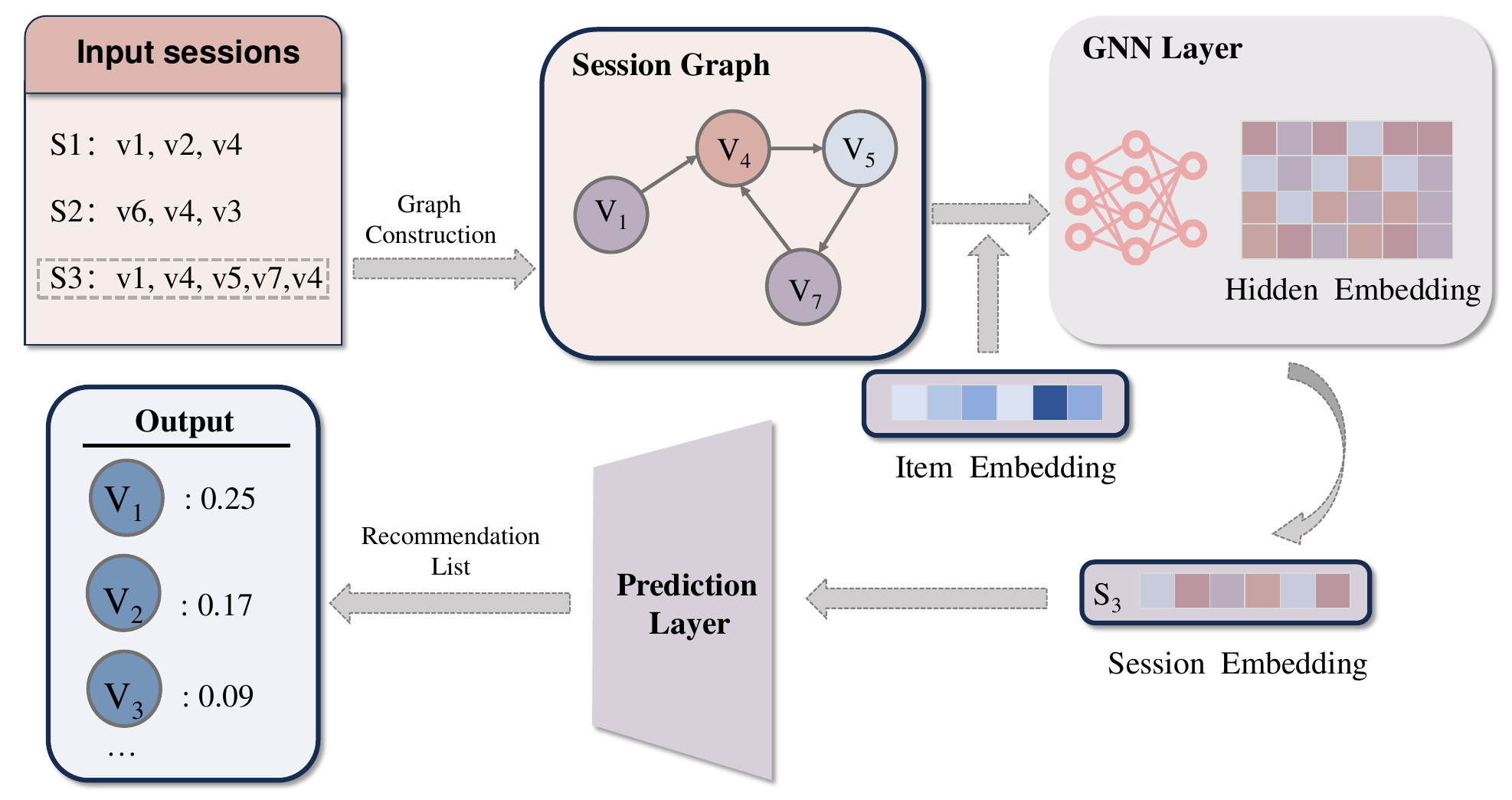}}
\caption{The GNN-based link prediction framework for a session-based recommendation system. Input sessions are transformed into session graphs to capture item transition patterns. A GNN layer aggregates neighborhood features into hidden embeddings, which are synthesized into a global session representation to predict the ranked recommendation list for the next item. \label{fig:GNN4SBR}}
\end{figure*}

Traditional recommendation systems rely on user identity and long-term behavior data for personalized recommendations \cite{sarwar-2001-item,koren2008factorization}. However, with the strengthening of privacy policies and the increase in anonymous users, these methods are no longer applicable. SBR have emerged \cite{li-2017-neural,wu-2019-session,zhang_beyond_2024}, which utilize only short-term behavior sequences within the user's current session to predict the next behavior. SBR predicts the next user interaction based on behavioral patterns within the current session, which can be seen as a special link prediction problem. Recently, Numerous advanced approaches transform user session behavior sequences into graph representations  \cite{wang-2020-global,qiu2020exploiting,pang-2022-heterogeneous,wang_incorporating_2023,zhang_bi-preference_2024,wang_exploiting_2024,yin_h3gnn_2024}, capitalizing on the inherent advantages of GNN for processing non-Euclidean data structures. The basic GNN-based method for SBR is shown as Fig. \ref{fig:GNN4SBR}. The inherent capability of GNNs to dynamically capture both intra-session item transition relationships and inter-session behavioral correlations has led to the successful application of GNN-based link prediction methodologies in recommendation systems, particularly for session-based scenarios where understanding temporal dependencies and latent connectivity patterns is crucial. BiPNet proposed by Zhang et al. \cite{zhang_bi-preference_2024} constructs a heterogeneous hypergraph, using users' behavioral features as different types of nodes, and captures high-order relationships through hyperedges, effectively capturing users' price preferences and interest preferences, successfully improving the performance of SBR. RNMSR \cite{wang_exploiting_2024} learns instance-level preferences through GNN, modeling users' repeat consumption behavior by combining instance-level and group-level user preferences, providing a new perspective for SBR. H3GNN \cite{yin_h3gnn_2024} integrates hierarchical GNN and directed graph aggregators to effectively capture both the high-order relationships between users and items and the sequential dependencies within sessions. LP-MRGNN \cite{wang_incorporating_2023} captures global item relationships across sessions by creating a multi-relational item graph that encompasses both target and auxiliary behaviors, which are encoded with GNN. Additionally, the model innovatively integrates the link prediction task into the loss function to assist in training, significantly enhancing recommendation performance. Moreover, xia et al. \cite{xia_efficient_2023} propose a novel model compression technique leveraging compositional encoding, which significantly accelerates the reconstruction of embeddings to enable faster local inference. 

\section{Challenges and open issues }
\label{sec6}

This survey has synthesized the extensive contributions of the existing literature, highlighting the remarkable advancements and inherent strengths of GNN-based link prediction in both techniques and applications. However, despite this progress, several critical weaknesses and research gaps persist, limiting the efficacy of existing models in increasingly sophisticated real-world scenarios. Specifically, challenges remain to address structural complexity, representation expressiveness, learning paradigm innovation, and the fusion of advanced techniques. To bridge these gaps and catalyze transformative future improvements, this section systematically delineates the pivotal challenges and open issues that demand urgent attention from the research community. By critically analyzing these limitations, we aim to provide a roadmap for developing next-generation GNN architectures that are more robust, scalable, and aligned with the multifaceted demands of advanced graph-based applications.

\subsection{Complex graphs}
While GNNs have demonstrated effectiveness in addressing link prediction challenges, existing methods predominantly focus on canonical graph structures, such as simple graphs and heterogeneous graphs. However, real-world networks exhibit multifaceted complexity, including multilayer graphs, hypergraphs, and dynamic graphs. Although recent efforts have leveraged attention mechanisms to incrementally adapt GNNs for certain complex scenarios, these solutions often remain superficial extensions of simple graph-based frameworks. Crucially, systematic GNN architectures tailored for advanced topologies (e.g., hypergraph, dynamic multilayer graph) remain underexplored. This gap necessitates a dedicated investigation into principled designs, such as neighborhood rough sets-based ensemble methods \cite{jiang2025neighborhood} and granular combination entropy \cite{jiang2026ensemble} for uncertainty modeling.

\subsection{Structural features}
GNN-based link prediction methods have become the standard in link prediction technology. Most of these methods follow a paradigm: using GNNs to learn individual node representations and aggregating the representations of both endpoints to predict the link state. However, this approach has limitations in capturing the complex relationships between node pairs. GNNs have been shown to be incapable of differentiating node pairs that contain isomorphic nodes \cite{zhang_labeling_2022}. This is primarily because GNNs tend to learn identical representations for isomorphic nodes, and the simple aggregation of node representations cannot distinguish these node pairs. Enhancing the expressiveness of GNNs in multi-node tasks such as link prediction may be an interesting and challenging research direction. This challenge echoes the fundamental goals of traditional graph mining \cite{rehman2012graph}, which focuses on extracting hidden knowledge and structural patterns from complex databases. In fact, link prediction can be viewed as an extension of structural pattern analysis, similar to how frequent subgraph mining \cite{rehman2014performance,rehman2018efficient,rehman2020online,rehman2024study} identifies recurring and significant sub-structures to characterize network behaviors. While preliminary attempts have been made in this field, as analyzed in Section 4, coupled approaches \cite{tan2023bring} suffer from heavy computational burdens, whereas decoupled methods \cite{chamberlain2023graph} exhibit limited expressiveness due to their reliance on handcrafted structural features. A critical open challenge lies in learning expressive structural link representations efficiently (ensuring isomorphic links share identical embeddings while distinguishing all non-isomorphic counterparts  \cite{zhang_labeling_2022}) which remains unresolved. 

\subsection{Self-Supervised learning}
Recent advancements \cite{liu_edge_2024,zhang2023line}  have witnessed a paradigm shift from traditional supervised learning to contrastive self-supervised frameworks for link prediction. Self-supervised learning (e.g., contrastive learning) has emerged as a powerful paradigm for learning representations without explicit labeling, which is particularly beneficial in the context of link prediction where labeled data can be scarce or expensive to obtain. However, existing contrastive learning approaches \cite{liu_edge_2024,zhang2023line} for link prediction have some limitations. Many contrastive learning methods require complex graph augmentations and large batch sizes to effectively learn discriminative representations, making them computationally expensive. Furthermore, although contrastive learning can learn node representations by contrasting positive and negative sample pairs, it still faces challenges when dealing with unseen node pairs or new graph structures. By addressing these limitations and building on the strengths of self-supervised learning, contrastive learning can become an even more powerful tool for link prediction, offering both high performance and scalability.

\subsection{Scalability and computational efficiency}
With the exponential growth of real-world networks, link prediction models encounter substantial computational and scalability bottlenecks. Conventional mitigation strategies for large-scale graphs, including subgraph sampling, sparse neural operations, and distributed training frameworks, alleviate computational complexity but often compromise prediction accuracy due to structural information loss and gradient instability. Moreover, a deep architectural bottleneck known as “over-smoothing” frequently emerges when scaling GNNs, causing node features to converge and lose the discriminative structural identity required for edge inference. Consequently, designing lightweight architectures and topology-preserving partitioning methods remains a critical hurdle for efficient link inference. Beyond these architectural optimizations, the advent of Graph Foundation Models (GFMs) \cite{liu2025graph} represents a paradigm shift for relational reasoning. Analogous to Large Language Models (LLMs), GFMs are designed to extract universal topological and relational patterns by pre-training on vast, heterogeneous graph datasets \cite{shi2024lecture,10.1145/3711896.3737410}. For link prediction, this transfer learning approach circumvents the prohibitive cost of training bespoke models from scratch for every new network. Furthermore, GFMs offer a unified interface that bridges fragmented relational tasks, such as link prediction, triplet classification, and relation extraction. Building upon this trend, multimodal graph learning has also emerged, which enriches link inference by fusing structural data with diverse modalities like text and images to overcome the inherent sparsity of real-world graphs.

\section{Conclusion}
\label{sec7}
Graph Neural Networks (GNNs) have driven remarkable progress in addressing the inherent challenges of link prediction, attracting significant attention due to their diverse real-world applications. In this survey, we present a comprehensive overview of the latest advancements from a dedicated GNN perspective. We systematically categorize the latest research into techniques and applications based on the inherent dichotomy between methodological innovation and scenario-driven demands. In terms of technology, we deeply analyzed four primary architectural paradigms, namely GCN-based, GAE-based, GAT-based, and GFormer-based models. From the application perspective, we focus on link prediction for KGs and recommendation systems. In addition, we examine the current challenges and discuss promising future directions, including complex graphs, structural features, self-supervised learning, and scalability and computational efficiency. In a word, we systematically categorize and analyze these advanced techniques and their applications, aiming to facilitate our comprehension and utilization of GNN-based link prediction.

\section*{Acknowledgements}
We thank the partial support funded by Basic Research Program of Jiangsu (No.BK20251653), the Fundamental Research Funds for the Central Universities (No.2025QN1023), and the National Natural Science Foundation of China Joint Fund Key Project (No.U25B20138).

\bibliography{Manuscript}

\end{document}